\pdfoutput=1

\documentclass[11pt]{article}

\usepackage[]{EMNLP2023}

\usepackage{times}
\usepackage{latexsym}

\usepackage[T1]{fontenc}

\usepackage[utf8]{inputenc}

\usepackage{microtype}

\usepackage{inconsolata}

\usepackage{times}
\usepackage{latexsym}

\usepackage[T1]{fontenc}

\usepackage[utf8]{inputenc}

\usepackage{microtype}

\usepackage{inconsolata}

\def\ourmethod{AT-MoE}
\usepackage{times}
\usepackage{soul}
\usepackage{url}
\usepackage[utf8]{inputenc}
\usepackage{graphicx}
\usepackage{amsmath, amssymb}
\usepackage{amsthm}
\usepackage{booktabs}
\usepackage{algorithm}
\usepackage{algorithmic}
\usepackage{multirow}
\usepackage{color}
\usepackage{bm}
\usepackage{bbm}
\newcommand{\ignore}[1]{}

\title{{\ourmethod}: Adaptive Task-planning Mixture of Experts via LoRA Approach}

\author{Xurui Li\textsuperscript{\rm 1}, Juanjuan Yao\textsuperscript{\rm 1}\thanks{\quad Corresponding authors.} \\
        Shanghai Mingpin Medical Data Technology Co., Ltd., China \\ }

\begin{document}
\maketitle
\begin{abstract}
The advent of Large Language Models (LLMs) has ushered in a new era of artificial intelligence, with the potential to transform various sectors through automation and insightful analysis. The Mixture of Experts (MoE) architecture has been proposed as a solution to enhance model performance in complex tasks. Yet, existing MoE models struggle with task-specific learning and interpretability, especially in fields like medicine where precision is critical. This paper introduces the Adaptive Task-planing Mixture of Experts ({\ourmethod}), an innovative architecture designed to address these limitations. We first train task-specific experts via LoRA approach to enhance problem-solving capabilities and interpretability in specialized areas. Subsequently, we introduce a layer-wise adaptive grouped routing module that optimizes module fusion based on complex task instructions, ensuring optimal task resolution. The grouped routing module first perform overall weight allocation from the dimension of the expert group, and then conduct local weight normalization adjustments within the group. This design maintains multi-dimensional balance, controllability, and interpretability, while facilitating task-specific fusion in response to complex instructions.

\end{abstract}
 
\section{Introduction}
In the ever-evolving realm of artificial intelligence, large language models (LLMs) have emerged as a formidable force. They have demonstrated remarkable progress across a diverse range of domain tasks, holding the potential to revolutionize numerous industries and fields by automating complex tasks and offering intelligent insights. Despite their significant success, existing LLMs encounter substantial challenges in specific domains. For example, in areas such as coding and mathematics that demand a high level of reasoning ability, existing LLMs struggle to provide satisfactory results~\cite{yang2024harnessing}. This is attributed in part to the complex nature of these domains, where accurate solutions hinge on a profound understanding of logical principles and problem-solving strategies. Moreover, pre-trained LLMs often lack specialized domain knowledge, as is the case in the medical field~\cite{liu2024moe}. The limited availability of medical domain corpus during pre-training constrains their ability to handle tasks related to diagnosis, treatment, and medical research. In the medical field, the complexity and diversity of data pose formidable challenges to diagnosis and treatment. The extensive medical information, including patient records, research papers, and clinical guidelines, calls for sophisticated models that can accurately analyze and interpret this data. Traditional methods often falter in handling the complexity and variability of medical data, leading to limitations in diagnosis accuracy and treatment effectiveness~\cite{yang2023large}.

The Mixture of Experts (MoE) architecture presents a promising solution to these challenges~\cite{zhou2022mixture}. By amalgamating the expertise of multiple models, MoE architectures can handle complex tasks more effectively than individual models. Currently, the mainstream MoE architecture in the large model field focuses on introducing a sparsely activated MoE layer based on a gating mechanism at the model level and replacing the feedforward component of the Transformer layer. This approach offers better model capacity and computational flexibility without increasing inference cost, making it ideal for LLMs.

However, in complex scenarios such as medical field, particularly those requiring expertise and explainability, the existing MoE architecture still face difficulties in achieving high-quality fusion learning effects. In the original report of Mixtral 8x7B, it can be found that no obvious patterns in the assignment of experts based on the topic can be observed~\cite{jiang2024mixtral}. The distribution of expert assignment is very similar for all layers for different tasks. It suggests that no distinct expert patterns have been clearly learned in this MoE architecture. To overcome this issue, we propose an Adaptive Task-planing Mixture of Experts ({\ourmethod}) architecture. This architecture is designed to harness the expertise in multiple fields while ensuring the credibility, controllability, and interpretability of the model. We first train several task-specific experts using Parameter-efficient fine-tuning (PEFT) technology such as LoRA~\cite{han2024parameter}, enabling our sub-networks to have better problem-solving capabilities and interpretability in specialized areas. Subsequently, we train an innovative layer-wise adaptive grouped routing module, allowing more efficient module fusion based on complex task instructions, thereby providing the optimal response for task resolution.

\begin{figure*}[!ht]
    \centering
\includegraphics[width=15cm]{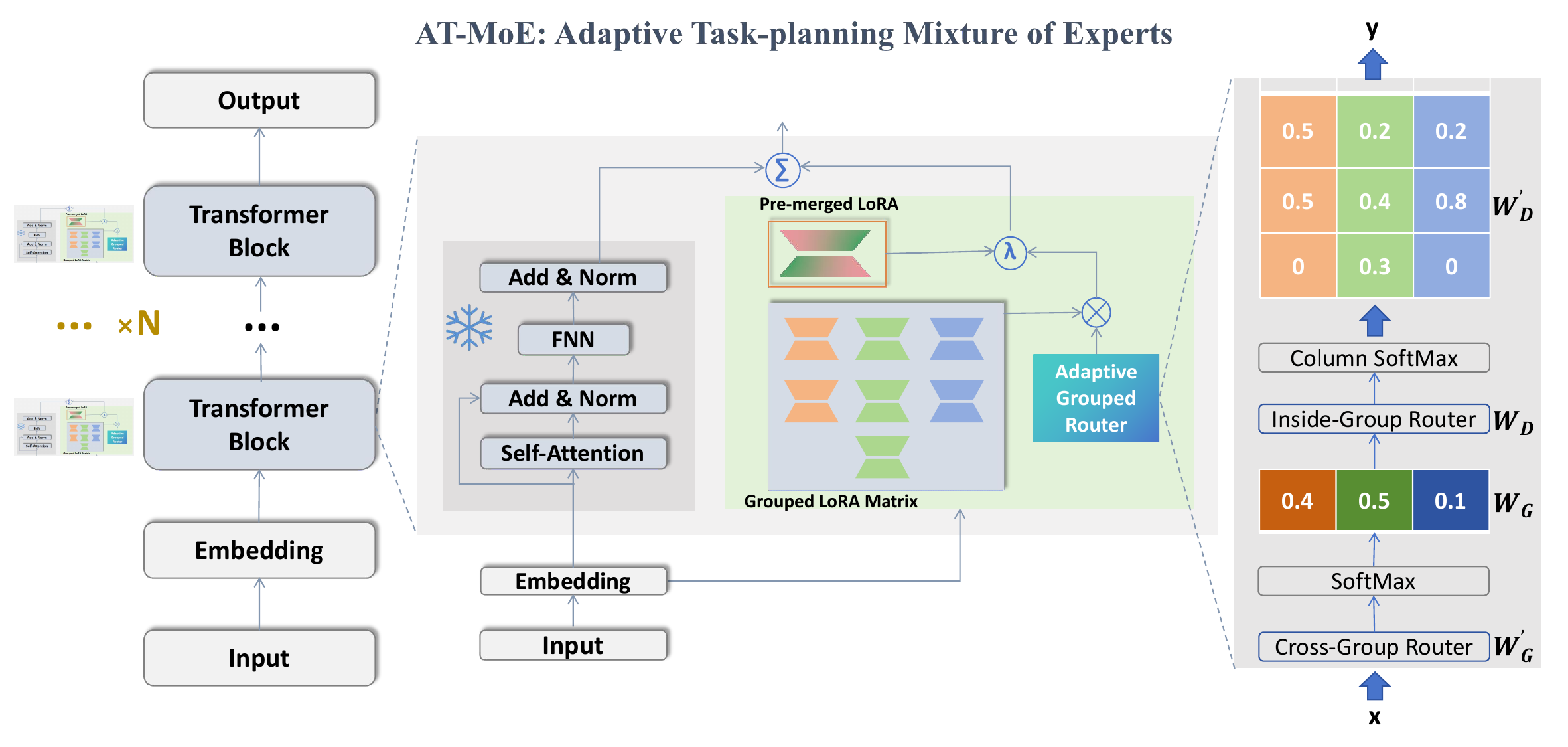}
    \caption{The Framework of {\ourmethod}. }
    \label{fig:model}
\end{figure*}


 

\section{{\ourmethod}: Methodology} 
We propose {\ourmethod} architecture to addresses the limitations of traditional MoE approaches when dealing with complex tasks, especially for scenarios where a single instruction contains multiple intents. The essence of the {\ourmethod} architecture revolves around its dynamic weighting allocation mechanism. This multi-module fusion approach enhances the system's capability to handle sophisticated and diverse challenges more effectively. Traditonal MoE lacks task-level training, so there is no expert corresponding to a specific task in the overall model architecture. The mixing and redundancy of knowledge hinders the degree of expert specialization. In comparison, we initially carried out the training of the expert network for diverse task scenarios. As a result, each expert network has a distinctively clear task domain attribute. To guarantee the efficiency of model training and inference, we employed the Parameter Efficient Fine-Tuning (PEFT) approach to train each expert sub-network. We emphasize that our {\ourmethod} is applicable to any PEFT method that introduces trainable modules throughout the model, such as LoRA, AdaLoRA, and (IA)3, etc. We use LoRA here due to its popularity and extensive usage. LoRA is a parameter- and compute-efficient finetuning method that incorporates trainable rank decomposition matrices into each layer of an LLM. It allows indirect training of dense layers with weight matrix, $W_{0} \in \mathbb R^{d \times k}$ by optimizing the rank-decomposition matrices by $W_{0}+ \Delta W = W_{0} +BA$, , where $B \in \mathbb R^{d \times r}$, $A \in \mathbb R^{r \times k}$, and the $r$ is much smaller than $d$ and $k$.



After all task-specialized LoRA modules have been trained, we train a gate module to determine which activations should be fed into the modules while keeping all the LoRA modules frozen. For complex task scenarios, we adopt a adaptive grouped routing strategy. We believe that for expert models, in accordance with different task types, weight allocation for grouping should be carried out first, followed by weight allocation within each group. It should be noted that the weight allocations within each group adhere to the distribution of $SoftMax$. Taking the medical scenario as an example, expert functions can be divided into three major categories. 1) The first category of experts primarily focuses on functional matters, such as case generation, pharmacy prescription, triage, and guidance. 2) The second category of experts mainly centers on domain knowledge-based issues. Each expert in this group represents professional knowledge in a specific field, such as surgery, radiology, and pathology. 3) The third category of experts is mainly responsible for style types. For instance, some experts only provide clear conclusions, while others offer additional opinions for reference.
Suppose a user inputs a question: ``\emph{Recently, I have weakness in all four limbs, easy fatigue, decreased appetite, and often feel full after eating. What disease do I have? Can you prescribe some traditional Chinese medicine for me to regulate?}'' According to this question, our routing module first allocates weights to the major groups. Since this question has a relatively clear demand for prescribing medicine, the weight of the functional-type major group is inherently higher. Within the functional expert group, the weight allocation focuses on ``\emph{diagnosis}'' and ``\emph{prescribing medicine.}'' In the weight allocation of the domain knowledge-based expert group, it emphasizes ``\emph{gastroenterology}'' and ``\emph{traditional Chinese medicine.}'' In the style expert group, ``\emph{rigorous}'' experts are preferred. It can be seen that through our designed adaptive grouped routing strategy, we can reasonably allocate expert weights for task scenarios and thereby provide the optimal comprehensive answer.

In specific implementation, we need to maintain two type of trainable matrices. The first matrix is the group routing vector $W_{G}$, with a size of $N_{dim}*N_{G}$, where $N_{dim}$ is the input's embedding size. Suppose for all the $N_{G}$ groups, the maximum number of sub-experts in each group is $N_{M}$. In specific training, we denote the adaptive group transformation as a function $F_{G}$(.). The detail for $F_{G}$(.) is described as follows:
we first map according to the input's embedding through $W_{G}$ to a cross-group weight vector $W_{G}$ of size $1*N_{G}$, and then we perform temperature-based $SoftMax$ on $M_{G}$ the obtain the $W^{'}_{G}$. The second matrix is the inside-group routing matrix $W_{D}$, with a size of $N_{G}*N_{M}$. For those groups with less than $N_{M}$ experts, We directly pad the remaining weights with negative infinity, so that they do not participate in the $SoftMax$ calculation. We then apply the temperature-based $SoftMax$ operation on the $W_{D}$ transformed matrix for each group column and achieve the $W^{'}_{D}$. Finally, we calculate the weighted sum from the dot product result between each task-specific LoRA matrix and the corresponding weight in $W^{'}_{D}$. In addition, we use the merged training dataset for all the tasks to train an pre-merged LoRA representing a general-purpose expert. The output for all the experts is expressed as:
\begin{equation}
 y_i = \left ( \lambda F_{G}(\overline{\bm W_{e}})  + (1-\lambda)W_{p} \right )x_i + W_{0}x_i
\end{equation}
,where $W_{0}$ is the pre-trained linear layer of the original LLM, $\overline{\bm W_{e}}$ is the matrix collection of the linear projection weight of the task-specific LoRA expert.
$W_{p}$ is the linear projection weight of the pre-merged LoRA expert. $\lambda$ is a balance parameter.

Recent studies have shown that higher layers learn more abstract and high-level information, and these features are used for downstream tasks~\cite{gao2024higher}. Instead of allocating different number of experts for different layers, we train different routing matrices for different transformer layers. Attention at different levels varies for different groups. It is supposed that higher layers focus more on formal or functional features, while lower layers pay more attention to foundational knowledge features. For a Transformer with $N_{T}$ blocks, we utilize a collection of $N_{T}$ trainable {\ourmethod} modules. The details for {\ourmethod} is shown in Fig.~\ref{fig:model}.

Note that during the training phase, {\ourmethod} predicts weights for multiple LoRAs. During the inference phase, {\ourmethod} can allocate the expert weights dynamically.



\section{Related Work}

\vspace{3pt}\noindent\textbf{Mixture of Experts}.
The concept of mixture of experts (MoE) has been extensively explored and advanced as demonstrated by subsequent studies~\cite{cai2024survey}. The emergence of sparsely-gated MoE, especially within the integration of transformer-based large language models, has infused new vitality into this technology that is three decades old~\cite{zhou2022mixture}. In the classical architecture of MoE, each MoE layer generally comprises a set of $N$ independent feed-forward networks (FFNs) that act as virtual experts within each Transformer block. Alongside this, there is a gating function which usually takes the form of a linear network with a $SoftMax$ activation function, signifying the weights assigned to the outputs of these experts. The independent feed-forward networks within each MoE layer bring diverse capabilities and perspectives, while the gating function acts as a coordinator, ensuring that the appropriate amount of influence is given to each expert's output based on the specific input characteristics. The dense MoE layer activates all expert networks in each iteration. Conversely, the sparsely-gated MoE layer is designed to activate only a chosen subset of experts during each forward pass~\cite{jiang2024mixtral}. This approach attains sparsity by computing a weighted sum of the outputs from merely the top-K experts instead of aggregating the outputs from all experts. However, it pays more attention to the optimization of performance while ignoring the effects of the model on different tasks. Some recent works have explored training MoE-style models on multitask mixtures. \citet{chronopoulou2023language} trains a system that routes each example among a set of experts and have demonstrated improved performance and mitigate the challenges associated with scaling instruction-tuned LLMs at scale~\cite{taskszadouripushing}. Alternatively, \citet{ponti2023combining} trains a skill matrix that learns to allocate a task to a set of skills, with each skill being a parameter-efficient module. However, all of these methods use the same set of data for the overall mixed training of both the router and the experts modules. The gating network and expert layers are jointly trained to minimize the overall loss function of the MoE model. The gating network learns to route each input to the most relevant expert layer(s), while the expert layers specialize in their assigned sub-tasks. Since these expert modules are not specifically trained on task-level data, there is actually no expert-level training for tasks, leading to lack of controllability and interpretability.

\vspace{3pt}\noindent\textbf{Integration of Expert Models}.
To address the controllability and interpretability problem of MoE architecture, several methods try to integrates the experts at the model. One direct approach is to choose which expert models to answer specific questions. For example, ZOOTER train a router to distribute queries among LLMs, guided by rankings from a distilled reward model, thus avoiding the activation of all LLMs for each query~\cite{lu2024routing}. MOLE train task-specific experts and then use task embeddings based on the diagonal of the Fisher information matrix to retrieve, average, and train modules from the top-K most similar tasks to a target task~\cite{wumixture}. Another approach aims to merge the capabilities of models trained on different tasks or domains into a single model. For example, \citet{chronopoulou2023language} merges separate task and language adapters to enable cross-lingual generalization. Recently, a Collaboration-of-Experts (CoE) method has been proposed to offer a more loose-coupled fusion pattern of different expert sub-networks~\cite{huang2024ccoe}. Instead of building the experts at the layer-level like MoE, CCoE framework integrates the experts at the model level. It clearly clarifies the functionality of each sub-network and gives a higher interoperability of understanding the working flow of the model. For the CoE method, each expert can be regarded as a complete language model (LLM) that has been trained for specific downstream tasks. Each task can only be handled by a single expert LLM selected through routing. However, in real-world situations, many instructions are composite tasks. A single task-oriented model often finds it difficult to achieve the optimal solution. Some practical applications have combined parameter-efficient modules with model fusion for task-level generalisation~\cite{liu2024moe,gou2023mixture,ponti2023combining,yang2024moral}. As an improvement on the linear arithmetic composition and reference tuning-based composition, MOLE introduce a learnable gating functions by utilize the outputs of multiple LoRAs at each layer to determine composition weights~\cite{wumixture}. Similarly, PHATGOOSE learns to route among specialized experts for zero-Shot generalization~\cite{muqeethlearning}.

\section{Conclusion}
In this paper, we introduce the {\ourmethod} architecture, which is designed to address the challenges faced by traditional MoE approaches in handling complex tasks that require expertise and explainability. The key innovation of {\ourmethod} lies in its dynamic weighting allocation mechanism by employing an adaptive grouped routing module for efficient module fusion based on complex task instructions, providing optimal responses for task resolution.

\bibliographystyle{acl_natbib}
\bibliography{anthology}

\end{document}